\documentclass{article}

\usepackage{iclr2027_conference,times}

\usepackage{amsmath}
\usepackage{amssymb}
\usepackage{array}
\usepackage{booktabs}
\usepackage{graphicx}
\usepackage{listings}
\usepackage{longtable}
\usepackage{tikz}
\usetikzlibrary{arrows.meta}
\usepackage{multirow}
\usepackage{pifont}
\usepackage{placeins}
\usepackage{xcolor}
\usepackage{hyperref}
\usepackage{url}
\usepackage{tabularx}

\newcommand{\method}{HSE}
\newcommand{\meanstd}[2]{#1_{\scriptscriptstyle\pm\,#2}}
\newcolumntype{L}[1]{>{\raggedright\arraybackslash}p{#1}}
\newcolumntype{C}[1]{>{\centering\arraybackslash}p{#1}}

\title{Hierarchical Skill Evolution for Coding Agents}

\author{
\begin{tabular}{c}
Jun Gao, Yun Peng, Changhai Zhou, Yuhan Zhu, Xiangyu Zeng, Shiyang Zhang \\
Yuhua Zhou, Luozheng Qin, Qian Qiao, Xiaoteng Ma, Baian Chen \\
Weiyi Shang, Zhen Ming Jack Jiang, Lizhi Liao, Xiaoxue Ren \\
\\[-0.3em]
\href{https://github.com/jungao1106/SKILL_EVO}
{\texttt{github.com/jungao1106/SKILL\_EVO}}
\end{tabular}
}

\iclrfinalcopy
\begin{document}

\maketitle

\begin{abstract}
Skills provide agents with reusable experience to guide future decisions without modifying the underlying model. 
However, coding trajectories are often entangled with task- and repository-specific contexts, 
making it challenging to distill procedural skills that generalize across diverse coding environments.
To address this challenge, we present \textbf{Hierarchical Skill Evolution (\method{})}, which progressively distills procedural knowledge from coding trajectories through hierarchical evidence aggregation.
\method{} first extracts task-level evidence from individual trajectories, then aggregates recurring evidence within the same repository to form repository-specific skill candidates, 
and finally consolidates consistent patterns across repositories into transferable skills across repositories. 
The evolution is driven by an external writer–evaluator loop that generates and assesses skills while keeping the underlying coding model frozen.
The loop externally updates the writer and evaluator policies to improve subsequent skill generation and assessment.
This hierarchy progressively filters task-specific details while preserving procedures that generalize across diverse coding environments. 
Successful trajectories yield reusable guidance for future tasks, while failed trajectories reveal recurring failure modes that guide avoidance and recovery.
We evaluate \method{} on SWE-bench Verified and DeepSWE, using GLM-5.2 across Pi and Claude Code harnesses and GPT-5.5 and Opus 4.8 with Pi.
Across both benchmarks, 10 skills evolved solely from SWE-Gym improve coding performance across all evaluated settings, demonstrating transfer across benchmarks, coding models, and agent harnesses.
We further evolve the skill library from target-benchmark trajectories without verifier feedback, relying solely on the learned evaluator for candidate assessment and achieving additional gains through benchmark-tailored  skills.

\end{abstract}

\section{Introduction}

Skills allow coding agents to reuse experience from previous tasks without modifying the underlying model
\citep{voyager2023,expel2024,awm2024,reasoningbank2026}. 
This is particularly attractive for repository-level coding, where every attempt produces a rich trajectory of searches, code inspection, edits, tests, and recovery actions \citep{sweagent2024,openhands2025}. 
Reusing these trajectories could prevent an agent from repeatedly rediscovering the same repair strategies. 
A trajectory, however, is not yet a reusable skill.
It is entangled with a particular issue, repository structure, file paths, and patch, whereas a useful skill should guide future decisions without assuming that these task-specific details transfer.

Existing forms of agent experience expose both sides of this problem. 
Raw trajectory retrieval preserves concrete evidence but can encourage agents to copy solutions that are specific to a previous task or repository \citep{sweexp2025}. 
SWESkillsBench provides broadly applicable domain knowledge, but this guidance may not tell an agent how to localize a particular bug, constrain a patch, validate the affected behavior, or recover after an incorrect edit \citep{sweskillsbench2026}. 
Methods that synthesize or optimize skills from trajectories demonstrate that experience can be converted into reusable instructions \citep{expel2024,awm2024,reasoningbank2026,trace2skill2024,skillopt2024,gepa2026}, but a central challenge remains: \emph{at what level should evidence be aggregated so that task-specific details are removed without also discarding the procedure that made the experience useful?}

We address this challenge with \textbf{Hierarchical Skill Evolution (\method{})}, which uses repository structure as an intermediate level to progressively aggregate task-level evidence into repository-level patterns and cross-repository procedural skills. This hierarchy filters task- and repository-specific details while preserving transferable procedures.
\method{} evolves successful evidence into procedural skills and recurring failure signatures into recovery skills for escaping unsuccessful repair directions. 
An external writer--evaluator loop jointly evolves the skill library and the policies governing skill generation and assessment \citep{reflexion2023,reasoningbank2026,protegi2023,gepa2026}, while the coding model remains frozen and receives only the evolved skills.

\begin{figure}[t]
\centering
\includegraphics[width=\linewidth]{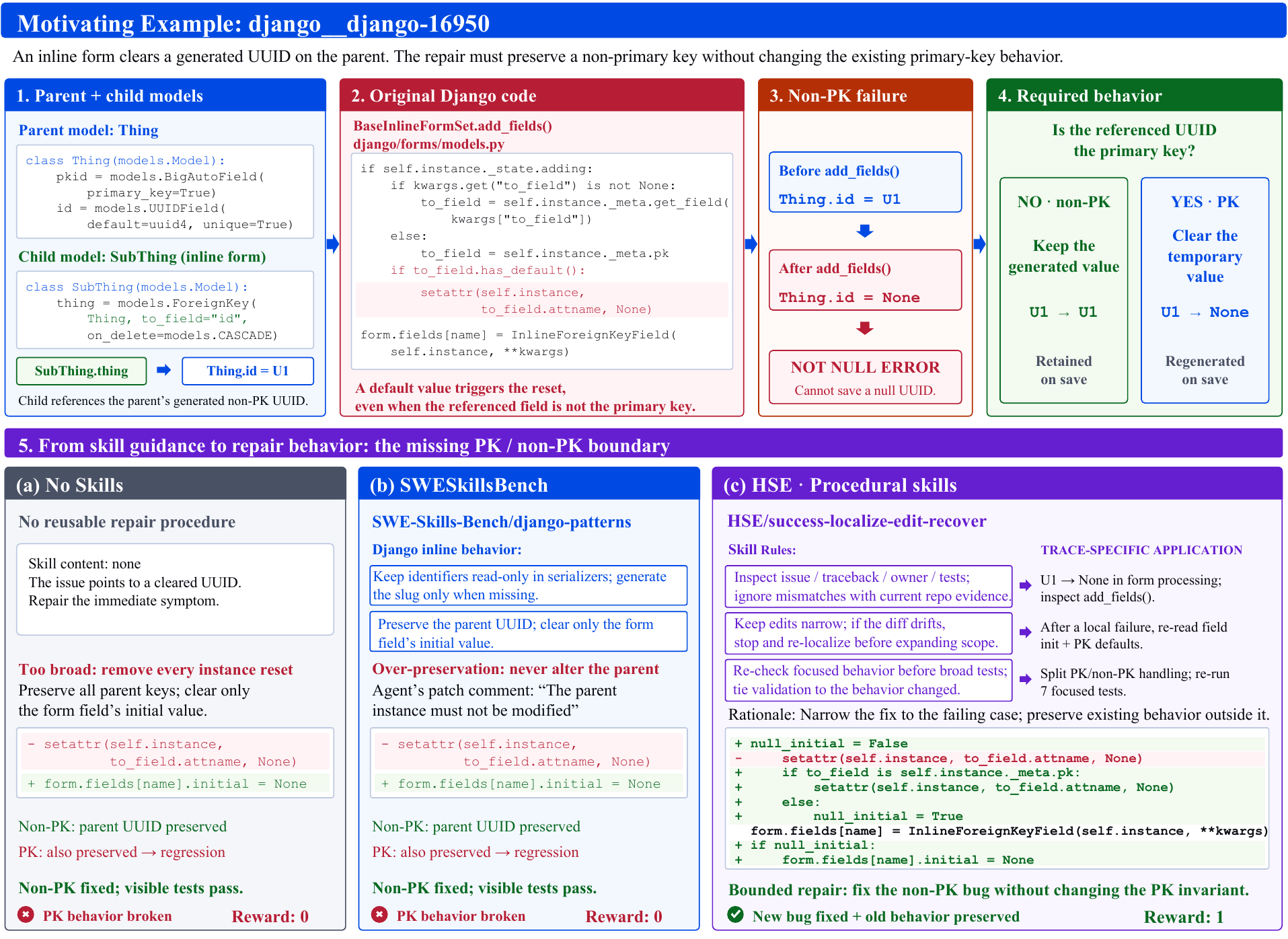}
\caption{Motivating example: \texttt{django\_\_django-16950}.
Panels~1--4 show the failure and the required primary-key boundary, and Panel~5 compares No Skills, SWESkillsBench, and HSE. 
Appendix~\ref{sec:motivation-study} provides the skill documents, patches, and trace evidence.}
\label{fig:motivation}
\end{figure}

Figure~\ref{fig:motivation} illustrates this distinction on \texttt{django\_\_django-16950}. 
A seemingly direct fix resolves the reported non-primary-key UUID failure but regresses the existing primary-key behavior, requiring the agent to discover the behavioral boundary rather than simply remove the offending operation. 
In the selected No Skills and SWESkillsBench traces, the agents produce the incomplete repair.
The HSE skill provides general guidance on minimal edits, focused validation, and re-localization.
In the selected HSE trace, the agent re-examines field initialization and primary-key behavior, identifies the PK/non-PK distinction from the current code and tests, and produces a conditional repair that preserves both behaviors.

We evaluate HSE with GLM-5.2 under Pi and Claude Code on SWE-bench Verified and DeepSWE, with additional cross-model evaluation using GPT-5.5 and Opus~4.8 under Pi.
Ten skills evolved solely on SWE-Gym consistently improve over No Skills and baselines across the target benchmarks and evaluated models, testing whether the learned procedures transfer beyond their source SWE-Gym environment.
We further study target-benchmark adaptation by transferring the learned writer and evaluator policies and updating the skill library from trajectories of the target task set. 
This experiment measures whether HSE can continue improving its skill library after exposure to target tasks without access to verifier feedback.

\begin{itemize}
    \item We formulate reusable skill evolution as a hierarchical process, progressing from task-level evidence through repository-conditioned patterns to reusable cross-repository procedures.
    
    \item We introduce a writer--evaluator framework that evolves successful and failed experience into procedural and recovery skills, while jointly updating the policies that govern subsequent skill construction and assessment.
    
    \item We evaluate frozen skill transfer, target-benchmark adaptation, component contributions, and cross-harness/model reuse across two coding benchmarks, two agent harnesses, and three underlying models.

\end{itemize}

\section{Related Work}


\paragraph{Learning from Agent Experience.}
ReAct interleaves reasoning and actions, Reflexion converts feedback into verbal reflection, and Self-Refine iteratively revises outputs using generated feedback~\citep{react2023,reflexion2023,selfrefine2023}. 
Voyager stores executable skills, and ExpeL extracts natural-language insights and retrieves past experience without parameter updates~\citep{voyager2023,expel2024}.
Agent Workflow Memory induces reusable routines from trajectories~\citep{awm2024}. 
ReasoningBank distills structured reasoning items from both successful and failed experience and retrieves them at test time~\citep{reasoningbank2026}. 
In coding, SWE-Exp retrieves prior trajectories under explicit leakage controls~\citep{sweexp2025}. 
These approaches motivate experience reuse, but do not perform HSE's task-to-repository-to-cross-repository aggregation of procedural evidence.

\paragraph{Skill and Prompt Evolution.}
Trace2Skill hierarchically merges trajectory-derived skill patches~\citep{trace2skill2024}, and SkillOpt aggregates success- and failure-derived edits into bounded skill updates \citep{skillopt2024}. 
AutoSkill extracts reusable skills from requirements, SkillOS trains a curator around a frozen executor, and MUSE-Autoskill manages and evaluates evolving skill packages~\citep{autoskill2023,skillos2025,museautoskill2026}. SkillRL combines general
and task-specific skills with policy learning~\citep{skillrl2026}. 
Automatic prompt optimization provides another route to improving external instructions.
ProTeGi edits prompts using textual feedback, whereas GEPA evolves prompts by reflecting on execution traces~\citep{protegi2023,gepa2026}. 
HSE instead updates the writer and evaluator policies that construct and screen a separately deployable skill library.


\section{Problem Formulation}
\label{sec:information-flow}

An issue-resolution task consists of an issue description, a repository snapshot, and an execution environment. 
A coding model interacts with the repository through an agent harness and produces a trajectory $\tau_i$ of tool calls and observations. 
HSE evolves reusable skills and the rules that construct them without updating the coding model.

During training, for each task $i$, an evidence extractor $\phi$ retains trace-visible execution signals associated with inspected code, edits, tests, and recovery actions, while $\psi$ summarizes the observed task outcome:
\[
  z_i=(e_i,y_i), \qquad e_i=\phi(\tau_i), \qquad
  y_i=\psi(\tau_i).
\]
These task records form L1 evidence. Let $\rho(i)$ denote the repository of
task $i$. HSE first groups records by repository,
\[
  \mathcal{E}_r=\{z_i:\rho(i)=r\},
\]
and applies a writer $\pi_w$ and evaluator $\pi_e$ with external rule sets
$\theta_w$ and $\theta_e$:
\[
  q_r=\pi_w(\mathcal{E}_r;\theta_w), \qquad
  a_r=\pi_e(q_r,\mathcal{E}_r;\theta_e).
\]
Here $q_r$ is a repository-conditioned candidate and
$a_r\in\{\text{accept},\text{revise},\text{reject}\}$ is its assessment.
This same-repository stage is L2.

L3 expands the support boundary across repositories. 
Accepted repository evidence is grouped by recurring procedural features to propose success skills, while repository failure evidence is grouped by recurring failure signatures to propose recovery skills. 
The writer converts each cross-repository group into a transferable candidate, and the evaluator checks its support, scope, and safety before promotion. 

The evolving state is
\[
  H_t=(K_t,\theta_w^t,\theta_e^t),
\]
where $K_t$ denotes the library of accepted procedural and recovery skills, while $\theta_w^t$ and $\theta_e^t$ govern candidate construction and assessment.
During training, a proposed state is evaluated on held-out validation tasks before it is accepted for the next iteration. 
The resulting transfer state has two consumers: the coding agent receives $K_t$, whereas the external writer $\pi_w$ and evaluator $\pi_e$ retain $\theta_w^t$ and $\theta_e^t$ to update their policies, respectively.

\section{Hierarchical Skill Evolution}
\label{sec:method-overview}

\begin{figure}[t]
\centering
\includegraphics[width=\linewidth]{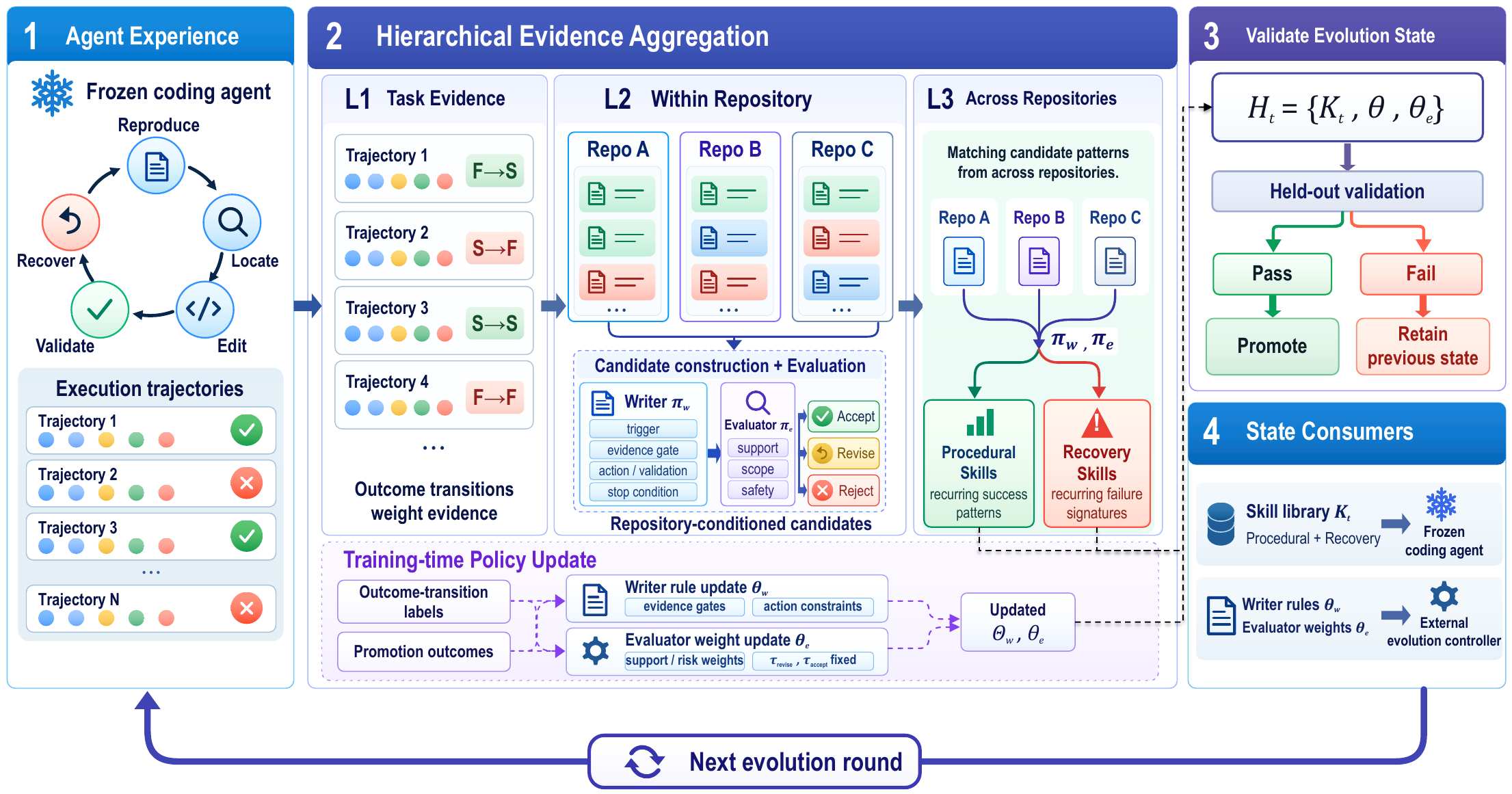}
\caption{HSE training pipeline. 
Task evidence is aggregated within repositories into repository-conditioned candidates and failure patterns, which are further consolidated across repositories into procedural and recovery skills.}
\label{fig:hse-pipeline}
\end{figure}

Figure~\ref{fig:hse-pipeline} summarizes HSE training as an iterative three-stage process.
First, a frozen coding agent produces execution trajectories and task outcomes. 
Second, HSE progressively expands the evidence boundary from tasks (L1), to repositories (L2), and then across repositories (L3).
In parallel, outcome-transition and promotion feedback update the external writer and evaluator rules.
Third, the proposed evolution state is validated and transferred to the next round.

\subsection{L1: Agent Rollouts and Task Evidence}

The frozen coding agent follows a recurring repair cycle with five stages: reproduce the problem, locate the responsible code, edit the implementation, validate the change, and recover when the current direction fails. 
Its tool interactions form an execution trajectory. During training, the task-level verifier supplies the observed outcome.

During training, L1 converts each trajectory into a task-level evidence record containing its five stage annotations and verifier outcome.
The five stage tags---reproduce, locate, edit, validate, and recover---associate files, code locations, patches, and tests with their roles in the issue-solving process. 
When a corresponding outcome from the previous evolution round is available, HSE compares it with the current verifier outcome to form an outcome transition label. 
This transition determines the strength of the task evidence.
Failed-to-successful and successful-to-failed transitions provide strong positive and strong negative evidence, respectively, whereas successful-to-successful and failed-to-failed transitions provide weak positive and weak negative evidence. 
When no paired outcome is available (in the first round), successful and unsuccessful trajectories are conservatively treated as weak evidence.
These labels determine how task evidence participates in subsequent aggregation. 
Positive evidence supports candidate skills construction and promotion, negative evidence prevents unsafe promotion, and recurring negative evidence contributes to recovery-skill construction.

\subsection{L2: Repository-Conditioned Candidates}

L2 aggregates labeled task evidence within each repository to identify recurring repository-level regularities.
Candidate support requires repeated positive evidence together with a recurring observable signal, such as a shared code path, edited path, or validation command. 
Strong negative evidence acts as a regression constraint and prevents unsafe promotion. 
This same-repository boundary preserves useful local structure.

The writer $\pi_w$ converts supported repository evidence into structured candidates.
Each candidate specifies a trigger, an evidence gate, recommended actions, validation steps, and a stop condition. 
The evaluator $\pi_e$ then assesses the candidate's empirical support, repetition across tasks, scope, and safety.
It accepts candidates that satisfy the promotion criteria, requests revision when additional specification or evidence is needed, and rejects candidates that are unsupported, unsafe, or overly task-specific.

L2 produces assessed repository-conditioned candidates and retains unresolved failure signatures for separate cross-repository aggregation at L3.

\subsection{L3: Transferable Skills}

L3 expands the support boundary from individual repositories to multiple repositories through two complementary aggregation paths.
Procedural skills are evolved from accepted repository-conditioned candidates, while recovery skills are distilled from recurring unresolved failure signatures.

\paragraph{Procedural Skills.}
Accepted repository evidence is grouped by recurring procedural features, such as localizing the narrowest supported owner, making a bounded edit, and validating the changed behavior. 
When the same procedural pattern is supported by multiple repositories, the writer removes repository-specific identifiers and proposes a transferable procedure. 
The evaluator verifies its cross-repository support before promotion.

\paragraph{Recovery Skills.}

Unresolved failure signatures are aggregated separately across repositories. 
Recurring signatures may capture behaviors such as localization drift, a collapsed state distinction, or validation that does not expose the changed behavior. 
The writer turns a supported cross-repository failure signature into a trigger and recovery procedure, while the evaluator checks whether the failure condition receives sufficient cross-repository support.
Recovery skills encode triggers and corrective procedures for recognizing and escaping recurring failure modes.

The accepted outputs of these two branches form the transferable skill library $K_t$. 
Cross-repository aggregation filters task- and repository-specific details while retaining procedures supported across diverse coding contexts.

\subsection{Writer and Evaluator Rule Updates}

During training, HSE periodically updates the external writer and evaluator rules that govern subsequent candidate construction and assessment.
Each update summarizes the task-evidence labels accumulated since the previous update together with the promotion outcomes observed during hierarchical aggregation.

HSE updates the writer rules to determine what evidence and actions may appear in subsequent candidates. 
Candidate construction remains grounded in trace-visible evidence and respects the evidence requirements of the corresponding aggregation level.
For example, accumulated unresolved evidence makes subsequent candidate construction more conservative, shifting the writer toward recovery and validation rather than direct editing.
Conversely, successful promotions reinforce evidence patterns that have survived the preceding gates for reuse in subsequent candidates.
These updates shape how the writer constructs future candidates, while candidate promotion remains determined by the evaluator.

The evaluator policy is parameterized by the weights assigned to the evidence features of each candidate. For a candidate $c$ with evidence features $f(c)$, the evaluator computes

$$
    s_{\theta_e}(c)=b + \theta_e^\top f(c),
$$

where $f(c)$ includes positive and negative support, task- or repository-level support, repeated-signal indicators, and other risk indicators. 
The resulting score is mapped to a decision using two fixed thresholds.
Candidates with scores above $\tau_{\mathrm{accept}}$ are promoted unless strong-negative evidence is present, those between $\tau_{\mathrm{revise}}$ and $\tau_{\mathrm{accept}}$ are retained for revision, and the remainder are rejected.

HSE evolves the evidence weights $\theta_e$ according to feedback accumulated after previous evaluation decisions, while keeping the decision thresholds $\tau_{\mathrm{revise}}$ and $\tau_{\mathrm{accept}}$ frozen. 
When a promoted candidate is subsequently associated with regression, HSE increases the penalties assigned to the negative or risk signals that contributed to its promotion. 
Conversely, when a revised or rejected candidate later accumulates consistent positive support, HSE increases the weights of the supporting signals that were previously undervalued. 
Strong-positive and strong-negative evidence induces larger updates than weak evidence, reflecting the greater informativeness of their outcome transitions.

\subsection{Held-Out Validation and Transfer}
\label{sec:training-loop}

During training, HSE treats the proposed skills and updated rules as one evolution state.
Before the state is accepted, the current skill library is evaluated on held-out tasks to determine whether performance improves, remains stable, or regresses. 
A validated state becomes the new version, and when validation rejects a proposed state, the current accepted version is retained.

The resulting transfer package contains success skills, recovery skills, and the updated writer/evaluator rules. 
The frozen coding agent receives the transferable skills, while the external controller receives the rules required to continue evolution in the next round.
New agent rollouts then provide L1 evidence for the next round, closing the loop in Figure~\ref{fig:hse-pipeline}.

\subsection{Training, Static Inference, and Target-Benchmark Adaptation}

During training, HSE uses SWE-Gym trajectories and verifier outcomes to construct the skill library and calibrate the writer and evaluator policies.
At static inference time, the learned state is frozen and only the skill library is provided to the coding agent.

For target-benchmark adaptation, HSE starts from the transferred skill library and fixed writer/evaluator policies, and updates only the skill library using newly collected target-task trajectories. 
Candidate updates are assessed by the learned evaluator without access to benchmark verifier outcomes or evaluation-time test results.

\subsection{Case: Conditional Repair in the HSE Trace}
\label{sec:method-case}

The HSE skill in Figure~\ref{fig:motivation}, panel~5(c),  instructs the agent to choose the narrowest owner supported by current evidence, keep the edit minimal, validate the affected behavior, and re-localize when a failed test contradicts the current repair direction.

In the selected trace, the agent first reproduces the UUID loss and inspects \texttt{add\_fields()}.
A PK-only reset fixes the reproducer but fails an existing initialization check. 
The agent then re-examines field initialization and the primary-key generation path.
The final patch resets default-valued primary keys before field construction, while preserving non-primary-key UUIDs on the parent instance and clearing only the form field's initial value. 
This conditional repair preserves both behaviors and passes the focused
tests. 
Appendix~\ref{sec:motivation-study} provides the skill listing, selected trace events, final patch, and verifier results.
\section{Evaluation}

We evaluate \method{} on transfer, target-benchmark adaptation, component contributions, and generalization across harnesses and models, with token-cost analysis referred to Appendix~\ref{sec:cost}.

\noindent\textbf{Setup.}
We evaluate \method{} on SWE-bench Verified (500 tasks) and DeepSWE (113 tasks) \citep{swebenchverified2024,deepswe2026}.
Unless otherwise specified, \method{} refers to the frozen skill library evolved on SWE-Gym and transferred to the target benchmark without further adaptation.
We report resolved rate, defined as the percentage of tasks with a successful verifier outcome.
All experiments use GLM-5.2 with Pi and Claude Code (CC) as coding-agent harnesses, except for the cross-model evaluation, which additionally considers GPT-5.5 and Opus~4.8 with Pi due to the compatibility constraint.
All reported differences in resolved rate are expressed in percentage points.

\noindent\textbf{Model Serving.}
We serve GLM-5.2 on two B300 nodes using prefill/decode (P/D) disaggregation.
The coding-agent harness manages agent interaction, including skill injection, while SGLang handles model inference.

\subsection{Static Inference}

Table~\ref{tab:main-results} compares HSE with No Skills, SWESkillsBench \citep{sweskillsbench2026}, GEPA \citep{gepa2026}, and ReasoningBank \citep{reasoningbank2026}. 
HSE achieves the highest mean resolved rate across all four benchmark--harness settings, improving over No Skills by 1.9--4.5 points on SWE-bench Verified and 5.6--7.7 points on DeepSWE. 
Notably, these gains are obtained by directly transferring the same skill library evolved on SWE-Gym, without any target-benchmark adaptation.

HSE also consistently improves over the strongest baseline on each benchmark. 
On SWE-bench Verified, it exceeds SWESkillsBench by 0.7--0.9 points, while on DeepSWE it exceeds GEPA by 2.1--3.5 points. 
In contrast, SWESkillsBench improves over No Skills on SWE-bench Verified but degrades performance on DeepSWE, suggesting that broadly applicable domain knowledge does not transfer uniformly across coding environments. 
HSE's consistent gains instead support the value of evolving procedural guidance that transfers across repositories and benchmarks. 
HSE also has lower between-run standard deviations than No Skills in all
four settings, indicating more consistent aggregate performance across the
three runs.

\begin{table}[!t]
\caption{Resolved rate (\%) of static skill methods on SWE-bench Verified
and DeepSWE with GLM-5.2. Pi and CC denote the Pi and Claude Code harnesses.
Entries report the mean over three runs (Mean@3), with standard deviations
shown as subscripts. HSE uses the frozen skill library evolved on SWE-Gym
without target-benchmark adaptation. Bold indicates the best mean in each row.}
\label{tab:main-results}
\begin{center}
\setlength{\tabcolsep}{2.8pt}
\begin{tabular}{@{}l l ccccc@{}}
\toprule
Benchmark & Harness & No Skills &  SWESkillsBench & GEPA & ReasoningBank & \method{} \\
\midrule
\multirow{2}{*}{SWE-bench Verified}
& Pi & $\meanstd{73.5}{1.10}$ & $\meanstd{74.7}{0.31}$ & $\meanstd{74.6}{0.40}$ & $\meanstd{72.8}{1.20}$ & $\mathbf{\meanstd{75.4}{0.20}}$ \\
& CC & $\meanstd{76.1}{1.30}$ & $\meanstd{79.7}{0.42}$& $\meanstd{78.9}{0.50}$& $\meanstd{76.6}{1.40}$& $\mathbf{\meanstd{80.6}{0.20}}$ \\
\cmidrule(l){2-7}

\multirow{2}{*}{DeepSWE}
& Pi& $\meanstd{43.1}{2.23}$& $\meanstd{34.5}{0.88}$& $\meanstd{46.6}{1.35}$& $\meanstd{44.0}{2.23}$& $\mathbf{\meanstd{48.7}{0.88}}$ \\
& CC& $\meanstd{44.8}{3.11}$& $\meanstd{38.6}{1.35}$& $\meanstd{49.0}{1.35}$& $\meanstd{46.0}{2.65}$& $\mathbf{\meanstd{52.5}{0.51}}$ \\
\bottomrule
\end{tabular}
\end{center}
\end{table}

\subsection{Target-Benchmark Adaptation}

Table~\ref{tab:online-evolution} evaluates target-benchmark adaptation.
Starting from the frozen HSE library transferred from SWE-Gym, each iteration collects target-task trajectories, updates the skill library with the fixed writer and evaluator, and reports performance after the evaluator gate.
This setting measures how effectively HSE continues to adapt its skill library as target-task experience accumulates.

Mean resolved rate increases at every reported gate across all four settings. 
By Gate~8, HSE improves from 75.4\% to 90.1\% with Pi and from 80.6\% to 93.9\% with CC on SWE-bench Verified, corresponding to gains of 14.7 and 13.3 percentage points. 
On DeepSWE, the corresponding scores rise from 48.7\% to 71.7\% and from 52.5\% to 75.8\%, yielding gains of 23.0 and 23.3 points. 
These results show that HSE can further refine its transferred skill library from target-task experience.

\begin{table}[!t]
\caption{Resolved rate (\%) of target-benchmark adaptation with GLM-5.2.
HSE denotes the transferred skill library before target-benchmark adaptation.
Gate 1--Gate 8 report the resolved rates achieved by the updated skill library after each evaluator gate.
Entries report Mean@3, with standard deviations shown as subscripts.}
\label{tab:online-evolution}
\begin{center}
\small
\resizebox{\linewidth}{!}{%
\begin{tabular}{@{}l l ccccccccc@{}}
\toprule
Benchmark & Harness & HSE & Gate 1 & Gate 2 & Gate 3 & Gate 4 & Gate 5 & Gate 6 & Gate 7 & Gate 8 \\
\midrule

\multirow{2}{*}{SWE-bench Verified}
& Pi
& $\meanstd{75.4}{0.20}$
& $\meanstd{81.1}{0.61}$
& $\meanstd{85.1}{0.70}$
& $\meanstd{86.1}{0.81}$
& $\meanstd{86.3}{0.50}$
& $\meanstd{87.6}{0.40}$
& $\meanstd{88.7}{0.31}$
& $\meanstd{89.4}{0.40}$
& $\meanstd{90.1}{0.31}$ \\

& CC
& $\meanstd{80.6}{0.20}$
& $\meanstd{86.4}{0.72}$
& $\meanstd{88.7}{1.01}$
& $\meanstd{88.9}{0.81}$
& $\meanstd{89.8}{0.92}$
& $\meanstd{91.7}{0.70}$
& $\meanstd{92.4}{0.40}$
& $\meanstd{93.0}{0.40}$
& $\meanstd{93.9}{0.46}$ \\

\cmidrule(l){2-11}

\multirow{2}{*}{DeepSWE}
& Pi
& $\meanstd{48.7}{0.88}$
& $\meanstd{55.8}{1.77}$
& $\meanstd{60.5}{1.35}$
& $\meanstd{61.7}{1.35}$
& $\meanstd{68.4}{2.23}$
& $\meanstd{69.6}{2.23}$
& $\meanstd{70.8}{0.88}$
& $\meanstd{71.1}{0.51}$
& $\meanstd{71.7}{0.88}$ \\

& CC
& $\meanstd{52.5}{0.51}$
& $\meanstd{55.8}{2.65}$
& $\meanstd{65.8}{2.23}$
& $\meanstd{70.8}{1.77}$
& $\meanstd{71.7}{0.88}$
& $\meanstd{72.9}{1.35}$
& $\meanstd{73.7}{1.02}$
& $\meanstd{74.9}{0.51}$
& $\meanstd{75.8}{0.51}$ \\

\bottomrule
\end{tabular}
}
\end{center}
\end{table}

\subsection{Component Analysis}

Table~\ref{tab:ablations} removes the cross-repository procedural skills or
recovery skills from the deployed HSE pack. Removing either component lowers
resolved rate in every setting, but their relative contributions reverse
across benchmarks. 
Procedural removal incurs the larger loss on SWE-bench Verified (1.2 and 3.6 points), while recovery removal incurs the larger loss on DeepSWE (5.3 and 5.6 points). 
We hypothesize that this difference reflects the greater task difficulty of DeepSWE, where agents are more likely to encounter failed attempts and therefore benefit more from recovery guidance.
For comparison, procedural removal costs 3.6 and 4.7 points on DeepSWE, and recovery removal costs 0.8 and 1.6 points on SWE-bench Verified.

\begin{table}[!t]
\caption{Resolved rate (\%) after removing HSE components. No Skills and HSE
repeat the main-table means without standard deviations. Parentheses show
the change from the complete HSE pack.}
\label{tab:ablations}
\begin{center}
\small
\resizebox{\linewidth}{!}{%
\begin{tabular}{@{}l l c c c c@{}}
\toprule
Benchmark & Harness & No Skills & HSE & w/o Procedural Skills & w/o Recovery Skills \\
\midrule
\multirow{2}{*}{SWE-bench Verified}
  & Pi & 73.5 & 75.4 & 74.2 ($-1.2$) & 74.6 ($-0.8$) \\
  & CC & 76.1 & 80.6 & 77.0 ($-3.6$) & 79.0 ($-1.6$) \\
\cmidrule(l){2-6}
\multirow{2}{*}{DeepSWE}
  & Pi & 43.1 & 48.7 & 45.1 ($-3.6$) & 43.4 ($-5.3$) \\
  & CC & 44.8 & 52.5 & 47.8 ($-4.7$) & 46.9 ($-5.6$) \\
\bottomrule
\end{tabular}
}
\end{center}
\end{table}

On DeepSWE, removing the two recovery skills leaves HSE only 0.3 points above
No Skills with Pi and 2.1 points above it with CC. Much of the full pack's
advantage on this benchmark is therefore lost even though the eight procedural
skills remain. This provides a more specific explanation for the main-result
pattern: guidance for revising an unsuccessful repair direction appears to
contribute beyond guidance for initiating a repair. The Django case in
Appendix~\ref{sec:motivation-study} illustrates such a revision process.

\begin{table}[t]
\caption{Ablation of hierarchical aggregation and controller updates.
All variants use the same SWE-Gym trajectories.
}
\label{tab:mechanism-ablation}
\centering
\small
\begin{tabular}{lcccc}
\toprule
& \multicolumn{2}{c}{SWE-bench Verified}
& \multicolumn{2}{c}{DeepSWE} \\
\cmidrule(lr){2-3}
\cmidrule(lr){4-5}
Method & Pi & CC & Pi & CC \\
\midrule
HSE
& 75.4 & 80.6 & 48.7 & 52.5 \\

w/o Hierarchical Aggregation
& 74.2 & 78.6 & 46.9 & 48.7 \\

w/o Controller Updates
& 74.8 & 79.2 & 46.9 & 50.4 \\
\bottomrule
\end{tabular}
\end{table}

Table~\ref{tab:mechanism-ablation} further isolates the contributions of hierarchical aggregation and controller updates in \method{}. 
Removing hierarchical aggregation randomly shuffles task trajectories across repositories during skill evolution, disrupting the original hierarchical aggregation while preserving the same trajectories and evidence budget. 
This tests whether the hierarchical organization of evidence itself contributes beyond simply exposing the skill evolution process to the same task experience.
This reduces performance by 1.2--2.0 points on SWE-bench Verified and 1.8--3.8 points on DeepSWE. 
Removing controller updates instead keeps the hierarchy but freezes the initial writer and evaluator rules throughout skill construction, resulting in losses of 0.6--1.4 and 1.8--2.1 points, respectively. 
These results indicate that both the hierarchical organization of evidence during skill evolution and the adaptive policies for constructing and assessing skill candidates contribute to the final transferred skill library.

\subsection{Generalization across Harnesses and Models}

Tables~\ref{tab:transfer} and~\ref{tab:additional-models} evaluate whether a
fixed HSE pack remains useful when the coding-agent harness or underlying
model changes.

\begin{table}[!t]
\caption{Resolved rate (\%) for fixed HSE packs across harnesses. No Skills
and in-harness HSE use the corresponding means from Table~\ref{tab:main-results}.
$\Delta_{\mathrm{NS}}$ and $\Delta_{\mathrm{in}}$ subtract the row's
transfer-harness No Skills and in-harness HSE scores from transfer-harness HSE,
respectively.}
\label{tab:transfer}
\begin{center}

\begin{tabular}{@{}l cc ccccc@{}}
\toprule
& \multicolumn{2}{c}{In-harness} & \multicolumn{5}{c}{Transfer harness} \\
\cmidrule(lr){2-3}\cmidrule(l){4-8}
Benchmark & Harness & HSE & Harness & No Skills & HSE & $\Delta_{\mathrm{NS}}$ & $\Delta_{\mathrm{in}}$ \\
\midrule
\multirow{2}{*}{SWE-bench Verified} & Pi & 75.4 & CC & 76.1 & 79.2 & $+3.1$ & $+3.8$ \\
& CC & 80.6 & Pi & 73.5 & 80.4 & $+6.9$ & $-0.2$ \\
\cmidrule(l){2-8}
\multirow{2}{*}{DeepSWE} & Pi & 48.7 & CC & 44.8 & 49.6 & $+4.8$ & $+0.9$ \\
& CC & 52.5 & Pi & 43.1 & 50.4 & $+7.3$ & $-2.1$ \\
\bottomrule
\end{tabular}
\end{center}
\end{table}

\begin{table}[!t]
\caption{Cross-model resolved rate (\%) with the Pi harness.}
\label{tab:additional-models}
\begin{center}
\small
\begin{tabular}{@{}l l c c@{}}
\toprule
Model & Method & SWE-bench Verified & DeepSWE \\
\midrule
\multirow{2}{*}{GPT-5.5} & No Skills & 82.6 & 67.2 \\
& HSE & 85.4 & 73.5 \\
\cmidrule(l){2-4}
\multirow{2}{*}{Opus 4.8} & No Skills & 88.6 & 59.3 \\
& HSE & 90.2 & 61.1 \\
\bottomrule
\end{tabular}
\end{center}
\end{table}

\FloatBarrier

Table~\ref{tab:transfer} shows that cross-harness deployment improves over the
transfer harness's No Skills result in all four settings, by 3.1--7.3 points.
Relative to the source harness's HSE score, changes range from $-2.1$ to
$+3.8$ points. The transferred packs remain useful across both harnesses.

Table~\ref{tab:additional-models} applies the same HSE pack to GPT-5.5 and
Opus~4.8 with Pi. HSE improves GPT-5.5 by 2.8 points on SWE-bench Verified
and 6.3 points on DeepSWE, and improves Opus~4.8 by 1.6 and 1.8 points,
respectively. Positive gains with both models support reuse of the same
procedural instructions beyond GLM-5.2.

\FloatBarrier

\section{Limitations and Conclusion}

The experiments span two benchmarks, two coding-agent harnesses, and three underlying models. 
Cross-model evaluation with GPT-5.5 and Opus 4.8 is conducted only under Pi, and the controller ablation evaluates the writer--evaluator update mechanism as a whole.

Across SWE-bench Verified and DeepSWE, the frozen HSE skill library improves over No Skills under both evaluated harnesses, and the same library remains useful when transferred to GPT-5.5 and Opus 4.8 under Pi.
Target-benchmark adaptation yields further gains as target-task experience accumulates, while the component and mechanism analyses show contributions from procedural and recovery skills, hierarchical aggregation, and controller updates.

\subsection*{AI Use Statement}
Generative AI assisted with manuscript drafting and polishing.

\subsection*{Reproducibility Statement}
Appendix~\ref{sec:motivation-study} provides the skill documents, exact patches, and selected trace events used in the motivating example.
Appendices~\ref{sec:implementation-details} and~\ref{sec:experiment-details}
specify the method implementation and experimental protocol.

\FloatBarrier
\label{page:main-end}
\clearpage
\bibliographystyle{iclr2027_conference}
\bibliography{references}

\clearpage
\appendix

\section{Cost of Transferable Skills}
\label{sec:cost}

Table~\ref{tab:token-cost} estimates context cost using word counts, avoiding
dependence on a provider's tokenizer. Generation counts model-produced text
and tool calls; prefilling counts the visible input to every model call,
including repeated history. The reported total is their sum, averaged over
one trace per task.

Relative to SWESkillsBench, HSE uses approximately 34\% and 39\% fewer total
words with Pi on SWE-bench Verified and DeepSWE, respectively, versus about
6\% and 3\% with CC. The step counts help interpret this difference: under
Pi, HSE uses fewer steps on both benchmarks, while under CC on DeepSWE it
uses slightly more steps (240.9 versus 237.3). A compact skill context can
therefore coexist with a longer interaction sequence. Moreover, prefilling
dominates the word totals in every condition, making repeated history an
important part of context cost beyond the initial skill injection.

HSE still uses more total words than No Skills in all four settings. Its
resolved-rate gains thus accompany additional context expenditure, while its
advantage over SWESkillsBench is associated with a smaller context footprint.

\begin{table}[!h]
\caption{Mean word-count cost using one trace per task. The reported average
injected skill context is 1.5K words for SWESkillsBench (49 skills) and 0.5K
words for HSE (eight procedural skills and two recovery skills).}
\label{tab:token-cost}
\begin{center}
\scriptsize
\setlength{\tabcolsep}{2.5pt}
\begin{tabular*}{0.98\linewidth}{@{\extracolsep{\fill}}llcccccccc@{}}
\toprule
& & \multicolumn{4}{c}{SWE-bench Verified} & \multicolumn{4}{c}{DeepSWE} \\
\cmidrule(lr){3-6}\cmidrule(lr){7-10}
Harness & Method & Generation & Prefilling & Steps & Total & Generation & Prefilling & Steps & Total \\
\midrule
\multirow{3}{*}{Pi}
& No Skills     & 2.1K & 312.5K & 52.4 & 314.7K & 8.3K & 2.4M & 141.9 & 2.4M \\
& SWESkillsBench & 3.4K & 578.8K & 79.2 & 582.2K & 14.8K & 4.6M & 223.6 & 4.6M \\
& HSE           & 2.5K & 382.6K & 58.4 & 385.1K & 9.0K  & 2.8M & 151.6 & 2.8M \\
\midrule
\multirow{3}{*}{CC}
& No Skills     & 3.4K & 450.6K & 59.2 & 454.0K & 12.3K & 4.5M & 184.2 & 4.5M \\
& SWESkillsBench & 6.2K & 698.7K & 79.8 & 704.9K & 17.3K & 6.9M & 237.3 & 6.9M \\
& HSE           & 3.5K & 659.1K & 74.0 & 662.6K & 16.8K & 6.7M & 240.9 & 6.7M \\
\bottomrule
\end{tabular*}
\end{center}
\end{table}

\section{Evidence Rules}
\label{sec:evidence-rules}

\paragraph{Task-level evidence construction.}
HSE constructs L1 evidence from trace-visible tool interactions.
A deterministic extractor records public execution signals, including touched
paths, edited paths, test commands, and the tool-call sequence. These signals,
together with skill resources actually accessed in the trace, are then
organized by a fixed model prompt into the five repair phases
\{\texttt{reproduce}, \texttt{locate}, \texttt{edit},
\texttt{validate}, \texttt{recover}\}.
Each phase record contains a trigger, actions, evidence to collect, and a stop
condition. Invalid model outputs fall back to a deterministic template-based
construction.

\paragraph{Aggregation and promotion.}
A repository-conditioned candidate must be supported by at least two positive task records and contain at least one repeated repository signal, defined by a recurring touched path, edited path, or test command. 
Candidates failing either condition are rejected before evaluator scoring. 
Candidates containing prohibited information, such as hidden patch content, task identifiers, or private sandbox paths, trigger the hard-safety filter and receive score zero.

Failure-derived candidates use cross-repository support as an evaluator feature. 
The support feature is activated at three supporting repositories during training.
Meeting the corresponding failure-support criterion contributes $0.16$ to the evaluator score.

For an eligible candidate $c$, the evaluator computes
\[
    s_{\theta_e}(c)=b + \theta_e^\top f(c).
\]
The decision thresholds are fixed throughout evolution:
\[
\tau_{\mathrm{revise}}=0.45,
\qquad
\tau_{\mathrm{accept}}=0.62.
\]
The final decision is
\[
a(c)=
\begin{cases}
\mathrm{accept}, &
s_{\theta_e}(c)\ge 0.62
\ \land\
\neg\mathrm{strong\_negative}(c),\\
\mathrm{revise}, &
0.45\le s_{\theta_e}(c)<0.62,\\
\mathrm{reject}, &
\textrm{otherwise}.
\end{cases}
\]

\paragraph{Writer and Evaluator Policy Updates}

The evaluator scores each candidate as
\[
s_{\theta_e}(c)=0.35+\theta_e^\top f(c),
\]
where $f(c)$ includes positive support, task support, repeated signals,
cross-repository support, event support, strong-negative evidence,
diagnostics, and risk indicators. The initial nonzero weights are
$0.08$ for positive support, $0.03$ for task support, $0.18$ for a
repeated signal, and $0.12$ when positive support reaches two tasks.
Failure-derived candidates additionally receive $0.10$ per supporting
repository, $0.01$ per supporting event, and $0.16$ when supported by
at least three repositories. Strong-negative, diagnostic, and risk
evidence contribute $-0.20$, $-0.04$, and $-0.08$, respectively.
The decision thresholds remain fixed at
$\tau_{\mathrm{revise}}=0.45$ and
$\tau_{\mathrm{accept}}=0.62$.

Every 40 new L1 task records, the evaluator updates the weights associated
with the observed evidence,
\[
\theta_{e,j}^{t+1}
=
\theta_{e,j}^{t}+\alpha(y)\Delta_j,
\qquad
\Delta_j\in\{-1,0,+1\},
\]
where strong evidence uses $\alpha=0.04$ and weak evidence uses
$\alpha=0.02$. $\Delta_j=+1$ increases and $\Delta_j=-1$ decreases
the corresponding evidence weight according to subsequent feedback.
Regression after promotion strengthens negative and risk-sensitive
weights, whereas subsequent positive support increases previously
undervalued supporting weights.

Writer updates are bounded rule changes over evidence gates, action
constraints, validation, and recovery. Each update introduces at most
three rule changes. 
\FloatBarrier
\section{Motivating Example: Skills and Selected Trajectories}
\label{sec:motivation-study}

This appendix documents the case in Figure~\ref{fig:motivation} using the
observed tool calls, source changes, and final evaluation.
The task is SWE-bench Verified's \texttt{django\_\_django-16950}:
16950 identifies the \href{https://github.com/django/django/pull/16950}{Django
repair pull request}, while the original issue is
\href{https://code.djangoproject.com/ticket/32210}{Django ticket 32210}.
We include the skill documents, source patches, and a chronological selection
of the consequential trace events.
The selection includes failed intermediate repairs and subsequent corrections
in all three conditions.

\lstdefinestyle{motivation-source}{
  basicstyle=\ttfamily\footnotesize,
  columns=fullflexible,
  keepspaces=true,
  breaklines=true,
  breakatwhitespace=false,
  showstringspaces=false,
  numbers=left,
  numberstyle=\tiny\color{gray},
  numbersep=6pt,
  xleftmargin=1.8em,
  frame=single,
  framerule=0.2pt,
  rulecolor=\color{gray},
  aboveskip=8pt,
  belowskip=8pt,
  tabsize=4,
  captionpos=t,
  literate={├}{{+}}1 {─}{{-}}1 {│}{{|}}1 {└}{{+}}1
}

\subsection{Task Context and the State Distinction}
\label{sec:motivation-context}

The parent has an auto-increment primary key, \texttt{pkid}, and a separate
unique, default-valued UUID field, \texttt{id}. The inline child references
\texttt{Thing.id}, not \texttt{Thing.pkid}. The following model definitions
are reproduced from the issue text in the No Skills trajectory with only
whitespace normalized; import and surrounding application setup are not shown.

\begin{lstlisting}[style=motivation-source,language=Python,caption={Parent and child definitions in the original task text.}]
class UUIDModel(models.Model):
    pkid = models.BigAutoField(primary_key=True, editable=False)
    id = models.UUIDField(default=uuid.uuid4, editable=False, unique=True)

    class Meta:
        abstract = True

class Thing(UUIDModel):
    name = models.CharField(max_length=191)

class SubThing(models.Model):
    name = models.CharField(max_length=191)
    thing = models.ForeignKey(
        'bugapp.Thing',
        to_field='id',
        on_delete=models.CASCADE,
        related_name='subthings',
    )
\end{lstlisting}

In Django's \texttt{BaseInlineFormSet.add\_fields()}, the implementation
clears a new parent's referenced field if that field has a default. In this
task, doing so destroys the non-PK UUID and saving the parent raises a
NOT NULL error. Default-valued primary keys require different handling:
their existing reset is supported by the PK-generation hook on save. The
source therefore needs to preserve the non-PK value without removing the
established PK reset. The instance value and a form field's \texttt{initial}
are separate state, clearing the latter does not clear the former.

\subsection{Skill Documents}
\label{sec:motivation-skill-exposure}

\paragraph{No Skills.}
This condition supplies no additional skill index.

\paragraph{SWESkillsBench.}
The run exposes 49 indexed entries, including \texttt{django-patterns}.
The SWESkillsBench Django document contains application layout, model and ORM
examples, DRF serializers, caching, signals, and middleware. Its concrete
examples include read-only serializer identifiers and generating a slug only
when the slug is absent. Neither example specifies how
\texttt{BaseInlineFormSet} should distinguish PK from non-PK UUID state.
The complete document is reproduced in Section~\ref{sec:full-sweskillsbench-skill}.

\paragraph{HSE Skills.}
H5 reads \texttt{success-localize-edit-recover}, listed in Section~\ref{sec:full-hse-skill}. 
The skill asks the coding agent to inspect current evidence, choose the narrowest supported owner path, keep edits minimal, re-check focused behavior, and re-localize if the diff drifts from the symptom.
It also prohibits copying a prior patch without fresh evidence or treating weak validation as success. 

\subsection{Trace Reading Convention}

N, S, and H prefix the original \texttt{step\_id} in No Skills, SWESkillsBench,
and HSE Skills, respectively. They are trace-local event identifiers, not
aligned time or compute units. Tables below summarize the consequential
actions and observations in their original order. Code and output listings
are verbatim excerpts unless an editorial omission is explicitly marked.

\subsection{No Skills: Local Tests Pass, but the Patch Removes a PK Invariant}
\label{sec:trace-no-skills}

\begin{longtable}{@{}L{0.09\linewidth} L{0.41\linewidth} L{0.43\linewidth}@{}}
\caption{Selected No Skills events, in their original order.}
\label{tab:trace-no-skills}\\
\toprule
Step & Recorded action & Observation and relevance \\
\midrule
\endfirsthead
\toprule Step & Recorded action & Observation and relevance \\\midrule
\endhead
N11 & Read \texttt{add\_fields()} at the referenced-field/default branch.
& The code executes \texttt{setattr(..., None)} on the parent instance. \\
N28 & Create and run a non-PK UUID/inline-form reproducer.
& The form and formset are valid, but saving raises
\texttt{IntegrityError: NOT NULL constraint failed: \_\_main\_\_\_thing.id}. \\
N56 & Run model-form and formset suites after an early repair.
& The recorded result is \texttt{FAILED (failures=4, errors=1)};
one visible failure checks that the inline field's \texttt{initial} is None. \\
N65 & Replace the early change with the original default-field clearing block.
& The edit succeeds. This is an actual correction after failure, not an
unrevised first guess. \\
N78 & Construct \texttt{InlineForeignKeyField} before the default-field branch
and replace instance clearing with \texttt{form.fields[name].initial = None}.
& The edit succeeds. No PK/non-PK branch is added; both parent key types
remain populated. \\
N91--N92 & Add an alternate-key save regression test and run the UUID suite.
& \texttt{Ran 7 tests in 0.011s}; \texttt{OK}. \\
N94 & Temporarily withdraw the source fix, run the added test, then restore it.
& The \texttt{assertIsNotNone} check on \texttt{parent.uuid} fails on the original code.
The output confirms restoration of the fix. \\
N99 & Run a wider selection of formset, admin, and form tests.
& \texttt{Ran 1583 tests in 31.578s}; \texttt{OK (skipped=51)}.
This count includes skipped tests. \\
\bottomrule
\end{longtable}

The final repair preserves the non-PK UUID, resolving the reported failure,
but also preserves default-valued PKs that the old branch reset. Its visible
tests do not detect this instance-level regression. Listing~\ref{lst:no-patch}
is the actual tool-returned unified patch at N78, including the movement of
field construction; it is not a schematic reconstruction.

\begin{lstlisting}[style=motivation-source,caption={No Skills source edit at N78: the returned unified patch.},label={lst:no-patch}]
--- /testbed/django/forms/models.py
+++ /testbed/django/forms/models.py
@@ -1169,20 +1169,23 @@
         # to correctly resolve the initial form value.
         if self.fk.remote_field.field_name != self.fk.remote_field.model._meta.pk.name:
             kwargs["to_field"] = self.fk.remote_field.field_name
 
+        form.fields[name] = InlineForeignKeyField(self.instance, **kwargs)
+
         # If we're adding a new object, ignore a parent's auto-generated key
-        # as it will be regenerated on the save request.
+        # as it will be regenerated on the save request. Set the inline form
+        # field's initial value to None without mutating the parent instance
+        # (which would discard a field default, e.g. a UUIDField's default,
+        # and cause the parent save to fail with a NOT NULL constraint).
         if self.instance._state.adding:
             if kwargs.get("to_field") is not None:
                 to_field = self.instance._meta.get_field(kwargs["to_field"])
             else:
                 to_field = self.instance._meta.pk
             if to_field.has_default():
-                setattr(self.instance, to_field.attname, None)
+                form.fields[name].initial = None
 
-        form.fields[name] = InlineForeignKeyField(self.instance, **kwargs)
-
     def get_unique_error_message(self, unique_check):
         unique_check = [field for field in unique_check if field != self.fk.name]
         return super().get_unique_error_message(unique_check)
 
\end{lstlisting}

\subsection{SWESkillsBench: Retries Occur, but the Same Core Edit Remains}
\label{sec:trace-sweskillsbench}

\begin{longtable}{@{}L{0.09\linewidth} L{0.41\linewidth} L{0.43\linewidth}@{}}
\caption{Selected SWESkillsBench events, in their original order.}
\label{tab:trace-sweskillsbench}\\
\toprule Step & Recorded action & Observation and relevance \\\midrule
\endfirsthead
\toprule Step & Recorded action & Observation and relevance \\\midrule
\endhead
S57 & Run an admin/inline-form reproducer and print parent UUID state.
& UUID is present after parent-form validation but None after formset
validation; saving raises a NOT NULL error. \\
S65 & Run the existing UUID suite after an early edit.
& Six tests run, four fail. Failures include expectations that the inline
field's \texttt{initial} be None. \\
S72 & Move field construction earlier and clear only its \texttt{initial}.
& The successful edit removes \texttt{setattr(..., None)} for both key types.
Its comment says the parent instance must not be modified. \\
S74 & Re-run the reproducer and existing UUID tests.
& The output reports \texttt{FORMSET SAVED}; all six existing tests pass. \\
S86--S87 & Add an alternate-key save regression test and run the UUID suite.
& Seven tests pass. The output explicitly marks the old default-PK
submission test \texttt{ok}. \\
S89 & Substitute the original source, run the new test, then restore the fix.
& Equality of \texttt{parent.uuid} and \texttt{default\_uuid} fails on the
original source. This checks the repair, not the effect of removing a skill. \\
S108 & Read the final \texttt{models.py} block again.
& The code still clears only \texttt{form.fields[name].initial} when the
referenced field has a default; there is no PK/non-PK distinction. \\
\bottomrule
\end{longtable}

S87's visible result is important because the final evaluation later fails
a strengthened version of the same old-behavior test. The following output
is a contiguous excerpt from S87:

\begin{lstlisting}[style=motivation-source,caption={SWESkillsBench's visible old-PK test passes at S87.}]
test_inlineformset_factory_ignores_default_pks_on_submit (model_formsets.test_uuid.InlineFormsetTests.test_inlineformset_factory_ignores_default_pks_on_submit)
#24377 - Inlines with a model field default should ignore that default ... ok
\end{lstlisting}

The two baseline patches have the same core executable change. They differ
in explanatory comments and added test code, so they are not byte-identical
full patches. Listing~\ref{lst:sweskillsbench-patch} preserves the actual SWESkillsBench
comment separately from the SWESkillsBench skill document.

\begin{lstlisting}[style=motivation-source,caption={SWESkillsBench source edit at S72: the returned unified patch.},label={lst:sweskillsbench-patch}]
--- /testbed/django/forms/models.py
+++ /testbed/django/forms/models.py
@@ -1169,20 +1169,23 @@
         # to correctly resolve the initial form value.
         if self.fk.remote_field.field_name != self.fk.remote_field.model._meta.pk.name:
             kwargs["to_field"] = self.fk.remote_field.field_name
 
+        form.fields[name] = InlineForeignKeyField(self.instance, **kwargs)
+
         # If we're adding a new object, ignore a parent's auto-generated key
-        # as it will be regenerated on the save request.
+        # as it will be regenerated on the save request. The parent instance
+        # must not be modified (e.g. its default value shouldn't be unset) as
+        # that could prevent it from being saved, so only the form field's
+        # initial value is reset.
         if self.instance._state.adding:
             if kwargs.get("to_field") is not None:
                 to_field = self.instance._meta.get_field(kwargs["to_field"])
             else:
                 to_field = self.instance._meta.pk
             if to_field.has_default():
-                setattr(self.instance, to_field.attname, None)
+                form.fields[name].initial = None
 
-        form.fields[name] = InlineForeignKeyField(self.instance, **kwargs)
-
     def get_unique_error_message(self, unique_check):
         unique_check = [field for field in unique_check if field != self.fk.name]
         return super().get_unique_error_message(unique_check)
 
\end{lstlisting}

\subsection{HSE Skills: Local Failure, Further Inspection, and Conditional Repair}
\label{sec:trace-hse}

\begin{longtable}{@{}L{0.09\linewidth} L{0.41\linewidth} L{0.43\linewidth}@{}}
\caption{Selected HSE Skills events, in their original order.}
\label{tab:trace-hse}\\
\toprule Step & Recorded action & Observation and relevance \\\midrule
\endfirsthead
\toprule Step & Recorded action & Observation and relevance \\\midrule
\endhead
H5 & Read the exported procedural skill's \texttt{SKILL.md}.
& The tool returns the procedural skill used in the repair. \\
H98 & Run a reproducer that prints parent field values before and after
formset validation.
& The generated UUID becomes None; saving fails. Both \texttt{getattr}
and \texttt{pre\_save} output for the non-PK field are None. \\
H105 & Read \texttt{add\_fields()} in \texttt{models.py}.
& The output exposes the unconditional default-field reset. \\
H107 & Search repository commit history with \texttt{git log -p -S}.
& The returned history shows the earlier expansion of clearing from PKs
to referenced fields with defaults. \\
H109--H111 & Restrict resetting to default PKs and re-run the reproducer.
& The one-test reproducer passes; the non-PK UUID is retained before
and after saving. The repair is not yet complete. \\
H121 & Run the existing UUID suite.
& Six tests run; the alternate-key test fails because the inline field's
\texttt{initial} is not None. This is a visible repository-test failure. \\
H123 & Read \texttt{InlineForeignKeyField}'s initialization code.
& The field copies its initial value from the referenced parent attribute,
or from the parent PK when no alternate field is specified. \\
H127 & Read \texttt{get\_pk\_value\_on\_save()}.
& The hook returns a field default when saving an instance with no PK.
This supports preserving the special PK reset path. \\
H140--H144 & Make successive edits and re-read the changed block.
& An intermediate version constructs the inline field before resetting the
PK; the constructor would therefore capture the old PK as its initial value. \\
H146 & Introduce \texttt{null\_initial} and reorder the final implementation.
& Reset PKs before field construction; for non-PKs, retain the instance
value and clear only the constructed field's initial value. \\
H148 & Run the existing UUID tests together with the reproducer.
& \texttt{Ran 7 tests in 0.015s}; \texttt{OK}. The reproducer prints the
same non-PK UUID before formset validation and after saving. \\
H159--H160 & Add the final alternate-key regression test and run the UUID suite.
& Seven tests pass. \\
H162 & Withdraw the source fix, run the added regression test, then restore it.
& The equality check against \texttt{original\_uuid} fails without the fix;
the output confirms restoration. \\
H170 & Inspect the final production-source diff.
& The final diff retains both the PK/non-PK branch and correct constructor
ordering, as reproduced in Listing~\ref{lst:hse-patch}. \\
H172 & Run broader validation.
& \texttt{Ran 1108 tests in 4.377s}; \texttt{OK (skipped=14)}. \\
\bottomrule
\end{longtable}

H121's failure concerns the form field's initial value, not the strengthened
instance-state assertion added by the final benchmark evaluator. Its command
and the consequential output lines are reproduced below; intervening
test-runner output is omitted.

\begin{lstlisting}[style=motivation-source,caption={H121: visible repository-test command and selected failure lines.}]
cd tests && python runtests.py model_formsets.test_uuid 2>&1 | tail -30

# Output excerpt:
    self.assertIsNone(formset.forms[0].fields["parent"].initial)
AssertionError: UUID('d08bf034-64b1-406d-8aa8-064bd87b2477') is not None
\end{lstlisting}

H123 and H127 supply the two relevant code facts. These are contiguous
excerpts from the returned source, not the coding agent's explanation:

\begin{lstlisting}[style=motivation-source,language=Python,caption={H123: the field initializes from parent-instance state.}]
            if self.to_field:
                kwargs["initial"] = getattr(self.parent_instance, self.to_field)
            else:
                kwargs["initial"] = self.parent_instance.pk
        kwargs["required"] = False
        super().__init__(*args, **kwargs)
\end{lstlisting}

\begin{lstlisting}[style=motivation-source,language=Python,caption={H127: the PK-on-save hook, exactly as read.}]
    def get_pk_value_on_save(self, instance):
        """
        Hook to generate new PK values on save. This method is called when
        saving instances with no primary key value set. If this method returns
        something else than None, then the returned value is used when saving
        the new instance.
        """
        if self.default:
            return self.get_default()
        return None
\end{lstlisting}

The final patch narrows the original reset instead of eliminating it for
all parent keys. It also preserves execution order: resetting a PK after
constructing the form field would leave that field initialized from the old
PK. The \texttt{null\_initial} flag delays only the non-PK form-field reset.

\begin{lstlisting}[style=motivation-source,caption={H170: complete returned production-source diff, including comments and ordering.},label={lst:hse-patch}]
diff --git a/django/forms/models.py b/django/forms/models.py
index 3fa04b821f..3f05a67d44 100644
--- a/django/forms/models.py
+++ b/django/forms/models.py
@@ -1172,15 +1172,29 @@ class BaseInlineFormSet(BaseModelFormSet):
 
         # If we're adding a new object, ignore a parent's auto-generated key
         # as it will be regenerated on the save request.
+        null_initial = False
         if self.instance._state.adding:
             if kwargs.get("to_field") is not None:
                 to_field = self.instance._meta.get_field(kwargs["to_field"])
             else:
                 to_field = self.instance._meta.pk
             if to_field.has_default():
-                setattr(self.instance, to_field.attname, None)
+                if to_field is self.instance._meta.pk:
+                    # The primary key is regenerated on save, so it can be
+                    # safely cleared on the instance (the InlineForeignKeyField
+                    # will then pick up None as its initial value).
+                    setattr(self.instance, to_field.attname, None)
+                else:
+                    # A non-primary key with a default (e.g. a unique UUIDField
+                    # referenced via to_field) isn't regenerated on save, so
+                    # the instance value must be preserved. Only the form
+                    # field's initial value is cleared to avoid triggering
+                    # validation on empty extra forms.
+                    null_initial = True
 
         form.fields[name] = InlineForeignKeyField(self.instance, **kwargs)
+        if null_initial:
+            form.fields[name].initial = None
 
     def get_unique_error_message(self, unique_check):
         unique_check = [field for field in unique_check if field != self.fk.name]
\end{lstlisting}

\subsection{Benchmark Verification and Added Assertions}
\label{sec:motivation-verifier}

The benchmark test patch adds the following two assertions after the existing
formset-validity check in
\texttt{test\_inlineformset\_factory\_ignores\_default\_pks\_on\_submit}:

\begin{lstlisting}[style=motivation-source,language=Python,caption={Evaluation-time assertions added to the existing PK test.}]
self.assertIsNone(formset.instance.uuid)
self.assertIsNone(formset.forms[0].instance.parent_id)
\end{lstlisting}

Here \texttt{uuid} is the primary key of the existing test model; it is not
the non-PK \texttt{id} of the issue's \texttt{Thing}. No Skills and SWESkillsBench
fail the first added assertion because their patches retain the parent PK.
HSE Skills passes the check. The three final evaluations use the same
assertions.

\begin{table}[htbp]
\centering
\small
\caption{Required checks and final outcomes for the selected attempts.
F2P/P2P refer to benchmark-required checks, not the total number of tests
executed by the verifier process.}
\label{tab:motivation-outcomes}
\begin{tabular}{@{}lcccc@{}}
\toprule
Condition & New-issue F2P & Existing-behavior P2P & Verifier tests & Reward \\
\midrule
No Skills & 1/1 & 2/3 & 6/7 pass & 0 \\
SWESkillsBench & 1/1 & 2/3 & 6/7 pass & 0 \\
HSE Skills & 1/1 & 3/3 & 7/7 pass & 1 \\
\bottomrule
\end{tabular}
\end{table}

\clearpage
\subsection{Complete SWESkillsBench Django Skill}
\label{sec:full-sweskillsbench-skill}

The following listing reproduces the complete \texttt{django-patterns}
skill used by the SWESkillsBench baseline, including its front matter, code
examples, and quick-reference table.

\begin{lstlisting}[style=motivation-source,caption={Complete \texttt{django-patterns/SKILL.md} used by SWESkillsBench.},label={lst:full-sweskillsbench-skill}]
---
name: django-patterns
description: Django architecture patterns, REST API design with DRF, ORM best practices, caching, signals, middleware, and production-grade Django apps.
---

# Django Development Patterns

Production-grade Django architecture patterns for scalable, maintainable applications.

## When to Activate

- Building Django web applications
- Designing Django REST Framework APIs
- Working with Django ORM and models
- Setting up Django project structure
- Implementing caching, signals, middleware

## Project Structure

### Recommended Layout

```
myproject/
├── config/
│   ├── __init__.py
│   ├── settings/
│   │   ├── __init__.py
│   │   ├── base.py          # Base settings
│   │   ├── development.py   # Dev settings
│   │   ├── production.py    # Production settings
│   │   └── test.py          # Test settings
│   ├── urls.py
│   ├── wsgi.py
│   └── asgi.py
├── manage.py
└── apps/
    ├── __init__.py
    ├── users/
    │   ├── __init__.py
    │   ├── models.py
    │   ├── views.py
    │   ├── serializers.py
    │   ├── urls.py
    │   ├── permissions.py
    │   ├── filters.py
    │   ├── services.py
    │   └── tests/
    └── products/
        └── ...
```

### Split Settings Pattern

```python
# config/settings/base.py
from pathlib import Path

BASE_DIR = Path(__file__).resolve().parent.parent.parent

SECRET_KEY = env('DJANGO_SECRET_KEY')
DEBUG = False
ALLOWED_HOSTS = []

INSTALLED_APPS = [
    'django.contrib.admin',
    'django.contrib.auth',
    'django.contrib.contenttypes',
    'django.contrib.sessions',
    'django.contrib.messages',
    'django.contrib.staticfiles',
    'rest_framework',
    'rest_framework.authtoken',
    'corsheaders',
    # Local apps
    'apps.users',
    'apps.products',
]

MIDDLEWARE = [
    'django.middleware.security.SecurityMiddleware',
    'whitenoise.middleware.WhiteNoiseMiddleware',
    'django.contrib.sessions.middleware.SessionMiddleware',
    'corsheaders.middleware.CorsMiddleware',
    'django.middleware.common.CommonMiddleware',
    'django.middleware.csrf.CsrfViewMiddleware',
    'django.contrib.auth.middleware.AuthenticationMiddleware',
    'django.contrib.messages.middleware.MessageMiddleware',
    'django.middleware.clickjacking.XFrameOptionsMiddleware',
]

ROOT_URLCONF = 'config.urls'
WSGI_APPLICATION = 'config.wsgi.application'

DATABASES = {
    'default': {
        'ENGINE': 'django.db.backends.postgresql',
        'NAME': env('DB_NAME'),
        'USER': env('DB_USER'),
        'PASSWORD': env('DB_PASSWORD'),
        'HOST': env('DB_HOST'),
        'PORT': env('DB_PORT', default='5432'),
    }
}

# config/settings/development.py
from .base import *

DEBUG = True
ALLOWED_HOSTS = ['localhost', '127.0.0.1']

DATABASES['default']['NAME'] = 'myproject_dev'

INSTALLED_APPS += ['debug_toolbar']

MIDDLEWARE += ['debug_toolbar.middleware.DebugToolbarMiddleware']

EMAIL_BACKEND = 'django.core.mail.backends.console.EmailBackend'

# config/settings/production.py
from .base import *

DEBUG = False
ALLOWED_HOSTS = env.list('ALLOWED_HOSTS')
SECURE_SSL_REDIRECT = True
SESSION_COOKIE_SECURE = True
CSRF_COOKIE_SECURE = True
SECURE_HSTS_SECONDS = 31536000
SECURE_HSTS_INCLUDE_SUBDOMAINS = True
SECURE_HSTS_PRELOAD = True

# Logging
LOGGING = {
    'version': 1,
    'disable_existing_loggers': False,
    'handlers': {
        'file': {
            'level': 'WARNING',
            'class': 'logging.FileHandler',
            'filename': '/var/log/django/django.log',
        },
    },
    'loggers': {
        'django': {
            'handlers': ['file'],
            'level': 'WARNING',
            'propagate': True,
        },
    },
}
```

## Model Design Patterns

### Model Best Practices

```python
from django.db import models
from django.contrib.auth.models import AbstractUser
from django.core.validators import MinValueValidator, MaxValueValidator

class User(AbstractUser):
    """Custom user model extending AbstractUser."""
    email = models.EmailField(unique=True)
    phone = models.CharField(max_length=20, blank=True)
    birth_date = models.DateField(null=True, blank=True)

    USERNAME_FIELD = 'email'
    REQUIRED_FIELDS = ['username']

    class Meta:
        db_table = 'users'
        verbose_name = 'user'
        verbose_name_plural = 'users'
        ordering = ['-date_joined']

    def __str__(self):
        return self.email

    def get_full_name(self):
        return f"{self.first_name} {self.last_name}".strip()

class Product(models.Model):
    """Product model with proper field configuration."""
    name = models.CharField(max_length=200)
    slug = models.SlugField(unique=True, max_length=250)
    description = models.TextField(blank=True)
    price = models.DecimalField(
        max_digits=10,
        decimal_places=2,
        validators=[MinValueValidator(0)]
    )
    stock = models.PositiveIntegerField(default=0)
    is_active = models.BooleanField(default=True)
    category = models.ForeignKey(
        'Category',
        on_delete=models.CASCADE,
        related_name='products'
    )
    tags = models.ManyToManyField('Tag', blank=True, related_name='products')
    created_at = models.DateTimeField(auto_now_add=True)
    updated_at = models.DateTimeField(auto_now=True)

    class Meta:
        db_table = 'products'
        ordering = ['-created_at']
        indexes = [
            models.Index(fields=['slug']),
            models.Index(fields=['-created_at']),
            models.Index(fields=['category', 'is_active']),
        ]
        constraints = [
            models.CheckConstraint(
                check=models.Q(price__gte=0),
                name='price_non_negative'
            )
        ]

    def __str__(self):
        return self.name

    def save(self, *args, **kwargs):
        if not self.slug:
            self.slug = slugify(self.name)
        super().save(*args, **kwargs)
```

### QuerySet Best Practices

```python
from django.db import models

class ProductQuerySet(models.QuerySet):
    """Custom QuerySet for Product model."""

    def active(self):
        """Return only active products."""
        return self.filter(is_active=True)

    def with_category(self):
        """Select related category to avoid N+1 queries."""
        return self.select_related('category')

    def with_tags(self):
        """Prefetch tags for many-to-many relationship."""
        return self.prefetch_related('tags')

    def in_stock(self):
        """Return products with stock > 0."""
        return self.filter(stock__gt=0)

    def search(self, query):
        """Search products by name or description."""
        return self.filter(
            models.Q(name__icontains=query) |
            models.Q(description__icontains=query)
        )

class Product(models.Model):
    # ... fields ...

    objects = ProductQuerySet.as_manager()  # Use custom QuerySet

# Usage
Product.objects.active().with_category().in_stock()
```

### Manager Methods

```python
class ProductManager(models.Manager):
    """Custom manager for complex queries."""

    def get_or_none(self, **kwargs):
        """Return object or None instead of DoesNotExist."""
        try:
            return self.get(**kwargs)
        except self.model.DoesNotExist:
            return None

    def create_with_tags(self, name, price, tag_names):
        """Create product with associated tags."""
        product = self.create(name=name, price=price)
        tags = [Tag.objects.get_or_create(name=name)[0] for name in tag_names]
        product.tags.set(tags)
        return product

    def bulk_update_stock(self, product_ids, quantity):
        """Bulk update stock for multiple products."""
        return self.filter(id__in=product_ids).update(stock=quantity)

# In model
class Product(models.Model):
    # ... fields ...
    custom = ProductManager()
```

## Django REST Framework Patterns

### Serializer Patterns

```python
from rest_framework import serializers
from django.contrib.auth.password_validation import validate_password
from .models import Product, User

class ProductSerializer(serializers.ModelSerializer):
    """Serializer for Product model."""

    category_name = serializers.CharField(source='category.name', read_only=True)
    average_rating = serializers.FloatField(read_only=True)
    discount_price = serializers.SerializerMethodField()

    class Meta:
        model = Product
        fields = [
            'id', 'name', 'slug', 'description', 'price',
            'discount_price', 'stock', 'category_name',
            'average_rating', 'created_at'
        ]
        read_only_fields = ['id', 'slug', 'created_at']

    def get_discount_price(self, obj):
        """Calculate discount price if applicable."""
        if hasattr(obj, 'discount') and obj.discount:
            return obj.price * (1 - obj.discount.percent / 100)
        return obj.price

    def validate_price(self, value):
        """Ensure price is non-negative."""
        if value < 0:
            raise serializers.ValidationError("Price cannot be negative.")
        return value

class ProductCreateSerializer(serializers.ModelSerializer):
    """Serializer for creating products."""

    class Meta:
        model = Product
        fields = ['name', 'description', 'price', 'stock', 'category']

    def validate(self, data):
        """Custom validation for multiple fields."""
        if data['price'] > 10000 and data['stock'] > 100:
            raise serializers.ValidationError(
                "Cannot have high-value products with large stock."
            )
        return data

class UserRegistrationSerializer(serializers.ModelSerializer):
    """Serializer for user registration."""

    password = serializers.CharField(
        write_only=True,
        required=True,
        validators=[validate_password],
        style={'input_type': 'password'}
    )
    password_confirm = serializers.CharField(write_only=True, style={'input_type': 'password'})

    class Meta:
        model = User
        fields = ['email', 'username', 'password', 'password_confirm']

    def validate(self, data):
        """Validate passwords match."""
        if data['password'] != data['password_confirm']:
            raise serializers.ValidationError({
                "password_confirm": "Password fields didn't match."
            })
        return data

    def create(self, validated_data):
        """Create user with hashed password."""
        validated_data.pop('password_confirm')
        password = validated_data.pop('password')
        user = User.objects.create(**validated_data)
        user.set_password(password)
        user.save()
        return user
```

### ViewSet Patterns

```python
from rest_framework import viewsets, status, filters
from rest_framework.decorators import action
from rest_framework.response import Response
from rest_framework.permissions import IsAuthenticated, IsAdminUser
from django_filters.rest_framework import DjangoFilterBackend
from .models import Product
from .serializers import ProductSerializer, ProductCreateSerializer
from .permissions import IsOwnerOrReadOnly
from .filters import ProductFilter
from .services import ProductService

class ProductViewSet(viewsets.ModelViewSet):
    """ViewSet for Product model."""

    queryset = Product.objects.select_related('category').prefetch_related('tags')
    permission_classes = [IsAuthenticated, IsOwnerOrReadOnly]
    filter_backends = [DjangoFilterBackend, filters.SearchFilter, filters.OrderingFilter]
    filterset_class = ProductFilter
    search_fields = ['name', 'description']
    ordering_fields = ['price', 'created_at', 'name']
    ordering = ['-created_at']

    def get_serializer_class(self):
        """Return appropriate serializer based on action."""
        if self.action == 'create':
            return ProductCreateSerializer
        return ProductSerializer

    def perform_create(self, serializer):
        """Save with user context."""
        serializer.save(created_by=self.request.user)

    @action(detail=False, methods=['get'])
    def featured(self, request):
        """Return featured products."""
        featured = self.queryset.filter(is_featured=True)[:10]
        serializer = self.get_serializer(featured, many=True)
        return Response(serializer.data)

    @action(detail=True, methods=['post'])
    def purchase(self, request, pk=None):
        """Purchase a product."""
        product = self.get_object()
        service = ProductService()
        result = service.purchase(product, request.user)
        return Response(result, status=status.HTTP_201_CREATED)

    @action(detail=False, methods=['get'], permission_classes=[IsAuthenticated])
    def my_products(self, request):
        """Return products created by current user."""
        products = self.queryset.filter(created_by=request.user)
        page = self.paginate_queryset(products)
        serializer = self.get_serializer(page, many=True)
        return self.get_paginated_response(serializer.data)
```

### Custom Actions

```python
from rest_framework.decorators import api_view, permission_classes
from rest_framework.permissions import IsAuthenticated
from rest_framework.response import Response

@api_view(['POST'])
@permission_classes([IsAuthenticated])
def add_to_cart(request):
    """Add product to user cart."""
    product_id = request.data.get('product_id')
    quantity = request.data.get('quantity', 1)

    try:
        product = Product.objects.get(id=product_id)
    except Product.DoesNotExist:
        return Response(
            {'error': 'Product not found'},
            status=status.HTTP_404_NOT_FOUND
        )

    cart, _ = Cart.objects.get_or_create(user=request.user)
    CartItem.objects.create(
        cart=cart,
        product=product,
        quantity=quantity
    )

    return Response({'message': 'Added to cart'}, status=status.HTTP_201_CREATED)
```

## Service Layer Pattern

```python
# apps/orders/services.py
from typing import Optional
from django.db import transaction
from .models import Order, OrderItem

class OrderService:
    """Service layer for order-related business logic."""

    @staticmethod
    @transaction.atomic
    def create_order(user, cart: Cart) -> Order:
        """Create order from cart."""
        order = Order.objects.create(
            user=user,
            total_price=cart.total_price
        )

        for item in cart.items.all():
            OrderItem.objects.create(
                order=order,
                product=item.product,
                quantity=item.quantity,
                price=item.product.price
            )

        # Clear cart
        cart.items.all().delete()

        return order

    @staticmethod
    def process_payment(order: Order, payment_data: dict) -> bool:
        """Process payment for order."""
        # Integration with payment gateway
        payment = PaymentGateway.charge(
            amount=order.total_price,
            token=payment_data['token']
        )

        if payment.success:
            order.status = Order.Status.PAID
            order.save()
            # Send confirmation email
            OrderService.send_confirmation_email(order)
            return True

        return False

    @staticmethod
    def send_confirmation_email(order: Order):
        """Send order confirmation email."""
        # Email sending logic
        pass
```

## Caching Strategies

### View-Level Caching

```python
from django.views.decorators.cache import cache_page
from django.utils.decorators import method_decorator

@method_decorator(cache_page(60 * 15), name='dispatch')  # 15 minutes
class ProductListView(generic.ListView):
    model = Product
    template_name = 'products/list.html'
    context_object_name = 'products'
```

### Template Fragment Caching

```django
{% load cache %}
{% cache 500 sidebar %}
    ... expensive sidebar content ...
{% endcache %}
```

### Low-Level Caching

```python
from django.core.cache import cache

def get_featured_products():
    """Get featured products with caching."""
    cache_key = 'featured_products'
    products = cache.get(cache_key)

    if products is None:
        products = list(Product.objects.filter(is_featured=True))
        cache.set(cache_key, products, timeout=60 * 15)  # 15 minutes

    return products
```

### QuerySet Caching

```python
from django.core.cache import cache

def get_popular_categories():
    cache_key = 'popular_categories'
    categories = cache.get(cache_key)

    if categories is None:
        categories = list(Category.objects.annotate(
            product_count=Count('products')
        ).filter(product_count__gt=10).order_by('-product_count')[:20])
        cache.set(cache_key, categories, timeout=60 * 60)  # 1 hour

    return categories
```

## Signals

### Signal Patterns

```python
# apps/users/signals.py
from django.db.models.signals import post_save
from django.dispatch import receiver
from django.contrib.auth import get_user_model
from .models import Profile

User = get_user_model()

@receiver(post_save, sender=User)
def create_user_profile(sender, instance, created, **kwargs):
    """Create profile when user is created."""
    if created:
        Profile.objects.create(user=instance)

@receiver(post_save, sender=User)
def save_user_profile(sender, instance, **kwargs):
    """Save profile when user is saved."""
    instance.profile.save()

# apps/users/apps.py
from django.apps import AppConfig

class UsersConfig(AppConfig):
    default_auto_field = 'django.db.models.BigAutoField'
    name = 'apps.users'

    def ready(self):
        """Import signals when app is ready."""
        import apps.users.signals
```

## Middleware

### Custom Middleware

```python
# middleware/active_user_middleware.py
import time
from django.utils.deprecation import MiddlewareMixin

class ActiveUserMiddleware(MiddlewareMixin):
    """Middleware to track active users."""

    def process_request(self, request):
        """Process incoming request."""
        if request.user.is_authenticated:
            # Update last active time
            request.user.last_active = timezone.now()
            request.user.save(update_fields=['last_active'])

class RequestLoggingMiddleware(MiddlewareMixin):
    """Middleware for logging requests."""

    def process_request(self, request):
        """Log request start time."""
        request.start_time = time.time()

    def process_response(self, request, response):
        """Log request duration."""
        if hasattr(request, 'start_time'):
            duration = time.time() - request.start_time
            logger.info(f'{request.method} {request.path} - {response.status_code} - {duration:.3f}s')
        return response
```

## Performance Optimization

### N+1 Query Prevention

```python
# Bad - N+1 queries
products = Product.objects.all()
for product in products:
    print(product.category.name)  # Separate query for each product

# Good - Single query with select_related
products = Product.objects.select_related('category').all()
for product in products:
    print(product.category.name)

# Good - Prefetch for many-to-many
products = Product.objects.prefetch_related('tags').all()
for product in products:
    for tag in product.tags.all():
        print(tag.name)
```

### Database Indexing

```python
class Product(models.Model):
    name = models.CharField(max_length=200, db_index=True)
    slug = models.SlugField(unique=True)
    category = models.ForeignKey('Category', on_delete=models.CASCADE)
    created_at = models.DateTimeField(auto_now_add=True)

    class Meta:
        indexes = [
            models.Index(fields=['name']),
            models.Index(fields=['-created_at']),
            models.Index(fields=['category', 'created_at']),
        ]
```

### Bulk Operations

```python
# Bulk create
Product.objects.bulk_create([
    Product(name=f'Product {i}', price=10.00)
    for i in range(1000)
])

# Bulk update
products = Product.objects.all()[:100]
for product in products:
    product.is_active = True
Product.objects.bulk_update(products, ['is_active'])

# Bulk delete
Product.objects.filter(stock=0).delete()
```

## Quick Reference

| Pattern | Description |
|---------|-------------|
| Split settings | Separate dev/prod/test settings |
| Custom QuerySet | Reusable query methods |
| Service Layer | Business logic separation |
| ViewSet | REST API endpoints |
| Serializer validation | Request/response transformation |
| select_related | Foreign key optimization |
| prefetch_related | Many-to-many optimization |
| Cache first | Cache expensive operations |
| Signals | Event-driven actions |
| Middleware | Request/response processing |

Remember: Django provides many shortcuts, but for production applications, structure and organization matter more than concise code. Build for maintainability.
\end{lstlisting}

\clearpage
\subsection{HSE Procedural Skill}
\label{sec:full-hse-skill}

The following listing reproduces the HSE skill read at H5, with the run
identifier and path lists omitted.

\begin{lstlisting}[style=motivation-source,caption={HSE \texttt{success-localize-edit-recover/SKILL.md} used at H5.},label={lst:full-hse-skill}]
---
name: success-localize-edit-recover
description: Training-distilled success pattern for localization, narrow edit, and recovery.
active: true
quality_score: 0.81
quality_tier: success_pattern
risk_flags: []
use_policy: evidence-gated
level: success_pattern
status: frozen
skill_type: "success_pattern"
support_count: 5
event_support_count: 11
positive_support_count: 37
support_bucket: "support_3"
pattern_key: "localize+edit+recover"
trigger_index: 129
---

# success-localize-edit-recover

## Status

- Status: `frozen`
- Source: training-only repo-level success evidence
- Support: `5` repos; `11` repo events; `37` verifier-positive task events
- Support repos: Project-MONAI__MONAI, getmoto__moto, pandas-dev__pandas, pydantic__pydantic, python__mypy

## Applicability

Use this skill when current task evidence calls for localization, narrow edit, and recovery. It is a procedural prior, not a repo-specific patch recipe.

## Evidence Gate

Use only after inspecting the current repository evidence. Ignore it if the issue, traceback, owner path, or test signal does not match the current task.

## Actions

1. Start from the smallest public symptom: issue text, traceback, failing assertion, or focused test.
2. Use repeated owner paths only as localization priors; confirm with current repository evidence.
3. Choose the narrowest owner path that is supported by current evidence before editing.
4. Keep edits minimal and re-check the focused behavior before broad validation.
5. If the current diff drifts away from the symptom, stop and re-localize before expanding scope.

## Do Not

1. Do not copy historical paths, tests, or patch shape without fresh current-task evidence.
2. Do not broaden the edit solely because this skill was retrieved.
3. Do not treat weak validation as success; tie validation to the changed behavior.

## Support Summary

- Pattern key: `localize+edit+recover`
- Common focused tests: none
- Associated failure signatures: localization-drift
\end{lstlisting}

\section{Implementation Details}
\label{sec:implementation-details}

HSE maintains the skill library and writer/evaluator policies outside the frozen
coding model. The coding agent receives the exported skills, and the external
policies manage their construction and assessment.

\subsection{Hierarchical State}

\begin{table}[!htbp]
\caption{Three levels of evidence aggregation in HSE.}
\label{tab:levels}
\begin{center}
\small
\begin{tabular}{@{}c L{0.27\linewidth} L{0.55\linewidth}@{}}
\toprule
Level & Object & Role \\
\midrule
L1 & Task evidence & Organizes trajectory evidence into reproduce, locate,
edit, validate, and recover stages. \\
L2 & Repository candidate & Summarizes procedures supported by recurring
evidence within one repository. \\
L3 & Transferable skill & Retains success or recovery procedures supported
across multiple repositories. \\
\bottomrule
\end{tabular}
\end{center}
\end{table}

The writer and evaluator are external policies for candidate construction
and assessment at L2 and L3. During training, L1 records include verifier
outcomes and the outcome-transition labels described in
Section~\ref{sec:method-overview}.

\subsection{Candidate Representation}

Repository candidates and transferable skills share the representation
\[
  c=(\text{trigger},\text{evidence},\text{actions},
     \text{validation},\text{stop condition}).
\]
The trigger identifies when the skill applies, and the evidence gate specifies what must be observed before acting. 
Actions define the repair procedure, validation checks the changed behavior, and the stop condition determines when to abandon or re-localize the current direction.

Repository-conditioned candidates may refer to local APIs, paths, and tests.
L3 distills these candidates into reusable procedural instructions. 
Procedural skills aggregate positively supported trajectories, and recovery skills aggregate recurring failure signatures and corrective actions across repositories.
The evaluator accepts, revises, or rejects candidates based on their support, scope, and safety.

\section{Experimental Details}
\label{sec:experiment-details}

\subsection{Training and Validation Split}
\label{sec:data-split}

From SWE-Gym's 2,438 tasks across 11 repositories \citep{swegym2025}, we sample 500 tasks approximately evenly across repositories, using deterministic, evenly spaced sampling within each repository. 
The 26 tasks from \texttt{bokeh/bokeh} form the validation set, and the remaining 474 tasks from ten repositories form the training set (approximately 95\%/5\%). 
The two sets share no repositories.

\subsection{Benchmarks and Evaluation}

SWE-bench Verified contains 500 tasks, and DeepSWE contains 113 tasks.
A task is resolved when its patch passes the benchmark verifier. 
All resolved rates are computed over the complete benchmark.

For static HSE evaluation, the transferred skill library is frozen and no target-benchmark trajectories are used for skill construction or updating.

For target-benchmark adaptation, trajectories from the target benchmark are used to update the skill library, and the resulting library is evaluated on the same complete target task set.
Target-task verifier outcomes are used only to compute the reported resolved rates and are not provided as feedback to the writer or evaluator. 
The transferred writer and evaluator policies remain fixed throughout target-benchmark adaptation.
We use a fixed eight-round adaptation horizon for all settings, with no early stopping.
Each round consists of a complete benchmark rollout followed by one writer--evaluator skill update, and we report the resulting library state after every round.

The primary experiments use GLM-5.2 with Pi and Claude Code (CC).
No Skills uses no additional skill library, and SWESkillsBench supplies the domain-skill baseline. 
Static HSE uses ten skills evolved on SWE-Gym, including eight success skills and two recovery skills.

\paragraph{Statistical Reporting.}
For a benchmark with $N$ tasks and $k_r$ resolved tasks in run $r$, the Mean@3 reported in Tables~\ref{tab:main-results} and~\ref{tab:online-evolution} is
\[
    \operatorname{Mean@3}
    =
    \frac{100}{3N}\sum_{r=1}^{3} k_r.
\]
Subscripts report the sample standard deviation of the three per-run resolved rates, using the conventional $R-1$ denominator with $R=3$.
The benchmark denominators are $N=500$ for SWE-bench Verified and $N=113$ for DeepSWE. 
Tables~\ref{tab:main-run-counts} and~\ref{tab:evolution-run-counts} report the underlying integer resolved counts for each run. 
No Skills and in-harness HSE reference values in the component and cross-harness tables use the corresponding means from Table~\ref{tab:main-results}.

\begin{table}[!htbp]
\caption{Per-run resolved counts underlying the static-skill results in Table~\ref{tab:main-results}.}
\label{tab:main-run-counts}
\begin{center}
\small
\begin{tabular}{@{}llrrrr@{}}
\toprule
Benchmark & Harness & Method & Run 1 & Run 2 & Run 3 \\
\midrule

\multirow{10}{*}{SWE-bench Verified}
& \multirow{5}{*}{Pi}
& No Skills       & 362 & 367 & 373 \\
& & SWESkillsBench & 372 & 373 & 375 \\
& & GEPA            & 371 & 373 & 375 \\
& & ReasoningBank   & 358 & 364 & 370 \\
& & HSE             & 376 & 377 & 378 \\
\cmidrule{2-6}
& \multirow{5}{*}{CC}
& No Skills       & 374 & 380 & 387 \\
& & SWESkillsBench & 396 & 399 & 400 \\
& & GEPA            & 392 & 394 & 397 \\
& & ReasoningBank   & 376 & 383 & 390 \\
& & HSE             & 402 & 403 & 404 \\
\midrule

\multirow{10}{*}{DeepSWE}
& \multirow{5}{*}{Pi}
& No Skills       & 46 & 49 & 51 \\
& & SWESkillsBench & 38 & 39 & 40 \\
& & GEPA            & 51 & 53 & 54 \\
& & ReasoningBank   & 47 & 50 & 52 \\
& & HSE             & 54 & 55 & 56 \\
\cmidrule{2-6}
& \multirow{5}{*}{CC}
& No Skills       & 47 & 51 & 54 \\
& & SWESkillsBench & 42 & 44 & 45 \\
& & GEPA            & 54 & 55 & 57 \\
& & ReasoningBank   & 49 & 52 & 55 \\
& & HSE             & 59 & 59 & 60 \\
\bottomrule
\end{tabular}
\end{center}
\end{table}

\begin{table}[!htbp]
\caption{Per-run resolved counts underlying the target-benchmark adaptation results in Table~\ref{tab:online-evolution}.}
\label{tab:evolution-run-counts}
\begin{center}
\small
\resizebox{\linewidth}{!}{%
\begin{tabular}{@{}llrrrrrrrrr@{}}
\toprule
Benchmark & Harness &
HSE & Gate 1 & Gate 2 & Gate 3 & Gate 4 &
Gate 5 & Gate 6 & Gate 7 & Gate 8 \\
\midrule

\multirow{3}{*}{SWE-bench Verified}
& Pi$^1$ & 376 & 402 & 422 & 426 & 429 & 436 & 442 & 445 & 449 \\
& Pi$^2$ & 377 & 406 & 425 & 431 & 431 & 438 & 443 & 447 & 450 \\
& Pi$^3$ & 378 & 408 & 429 & 434 & 434 & 440 & 445 & 449 & 452 \\
\cmidrule{2-11}
& CC$^1$ & 402 & 428 & 439 & 440 & 444 & 455 & 460 & 463 & 468 \\
& CC$^2$ & 403 & 433 & 443 & 445 & 450 & 458 & 462 & 465 & 468 \\
& CC$^3$ & 404 & 435 & 449 & 448 & 453 & 462 & 464 & 467 & 472 \\
\midrule

\multirow{3}{*}{DeepSWE}
& Pi$^1$ & 54 & 61 & 67 & 68 & 75 & 76 & 79 & 80 & 80 \\
& Pi$^2$ & 55 & 63 & 68 & 70 & 77 & 79 & 80 & 80 & 81 \\
& Pi$^3$ & 56 & 65 & 70 & 71 & 80 & 81 & 81 & 81 & 82 \\
\cmidrule{2-11}
& CC$^1$ & 59 & 60 & 72 & 78 & 80 & 81 & 82 & 84 & 85 \\
& CC$^2$ & 59 & 63 & 74 & 80 & 81 & 82 & 84 & 85 & 86 \\
& CC$^3$ & 60 & 66 & 77 & 82 & 82 & 84 & 84 & 85 & 86 \\
\bottomrule
\end{tabular}%
}
\end{center}
\end{table}

\end{document}